%% file: emnlp2021.tex
\pdfoutput=1
\documentclass[11pt]{article}

\usepackage{emnlp2021}
\usepackage{times}
\usepackage{latexsym}

\usepackage[T1]{fontenc}
\usepackage[utf8]{inputenc}
\usepackage{float}
\usepackage{multirow}

\usepackage{microtype}
\usepackage{graphicx}
\usepackage{multicol}
\usepackage{CJKutf8}
\title{Cross-lingual Intermediate Fine-tuning improves Dialogue State Tracking}

\author{Nikita Moghe \and Mark Steedman \and Alexandra Birch \\
        School of Informatics, University of Edinburgh \\
        \texttt{\{nikita.moghe, a.birch\}@ed.ac.uk} , \texttt{steedman@inf.ed.ac.uk}
        }

\begin{document}
\maketitle
\begin{abstract}
Recent progress in task-oriented neural dialogue systems is largely focused on a handful of languages, as annotation of training data is tedious and expensive.  Machine translation has been used to make systems multilingual, but this can introduce a pipeline of errors. Another promising solution is using cross-lingual transfer learning through pretrained multilingual models. Existing methods train multilingual models with additional code-mixed task data or refine the cross-lingual representations through parallel ontologies. In this work, we enhance the transfer learning process by intermediate fine-tuning of pretrained multilingual models, where the multilingual models are fine-tuned with different but related data and/or tasks. Specifically, we use parallel and conversational movie subtitles datasets to design cross-lingual intermediate tasks suitable for downstream dialogue tasks. We use only 200K lines of parallel data for intermediate fine-tuning which is already available for 1782 language pairs.
We test our approach on the cross-lingual dialogue state tracking task for the parallel MultiWoZ (English$\rightarrow$Chinese, Chinese$\rightarrow$English) and Multilingual WoZ (English$\rightarrow$German, English$\rightarrow$Italian) datasets. We achieve impressive improvements (> 20\% on joint goal accuracy) on the parallel MultiWoZ dataset and the  Multilingual WoZ dataset over the vanilla baseline with only 10\% of the target language task data and zero-shot setup respectively.

\end{abstract}

\section{Introduction}
In recent years, task-oriented dialogue systems have achieved remarkable success by leveraging huge amounts of labelled data. This technology is thus limited to a handful of languages as collecting and annotating training dialogue data for different languages is expensive and requires supervision from native speakers \citep{Chen2018-yy}.

To avoid having to create large annotated datasets for every new language, recent works focus on transfer learning methods which use neural machine translation systems \citep{schuster-etal-2019-cross-lingual}, code-mixed data augmentation \citep{DBLP:conf/aaai/LiuWLXF20, DBLP:conf/ijcai/QinN0C20} or large multilingual models \citep{lin2021empirical}. Neural machine translation models incur additional overhead of training on millions of parallel sentences that may not be available for all language pairs. Code-mixed data augmentation methods involve replacing individual words from the source language with the target language by using parallel word pairs found in a dictionary. However, a simple synonym replacement may not be sufficient as the tasks become complicated. In this paper, we focus on transfer learning via large multilingual models, which will allow us to extend models to languages with limited labelled training data.  

\begin{figure*}[h]
    \centering
    \includegraphics[width=0.8\linewidth, height=7.2cm]{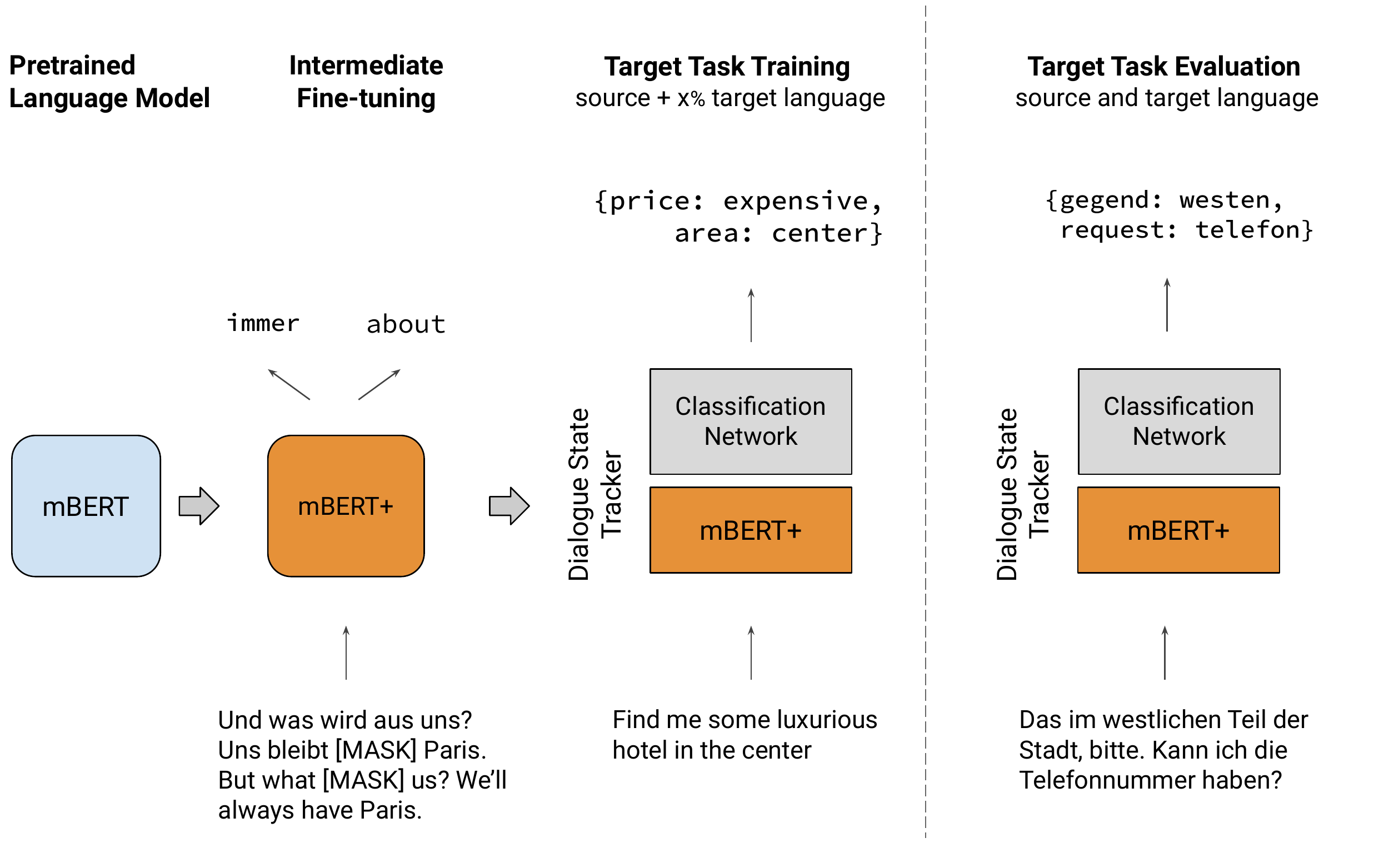}
    \caption{Pipeline of our work. A pretrained language model is fine-tuned with the task of predicting masked words on parallel movie subtitles data. A dialogue state tracker is then trained with this new multilingual model and evaluated for cross-lingual dialogue state tracking  }
    \label{fig:pipeline}
\end{figure*}

In techniques that use multilingual models, a task-specific architecture uses this pretrained model as one of its components and then is trained with task data from a high resource language (See Fig. \ref{fig:pipeline}). It is then evaluated directly or with some labelled examples in a different language. The use of intermediate fine-tuning, which is fine-tuning a large language model with a different but related data/or task and then fine-tuning it for the target task has shown considerable improvements for both monolingual and cross-lingual natural language understanding tasks \citep{gururangan-etal-2020-dont, phang-etal-2020-english}. But, it is relatively under-explored for multilingual dialogue systems. 

In this work, we demonstrate the effectiveness of using cross-lingual intermediate fine-tuning of multilingual pretrained models to facilitate the development of multilingual conversation systems. Specifically, we look at cross-lingual dialogue state tracking tasks, as they are an indispensable part of task-oriented dialogue systems. In this task, a model needs to map the user's goals and intents in a given conversation to a set of slots and values - known as a ``dialogue state'' based on a pre-defined ontology. Our intermediate tasks are based on interaction between the source and target languages and interaction between the dialogue history and response. These tasks involve the prediction of missing words in different conversational settings. These include monolingual conversations, concatenated parallel bilingual conversations, and cross-lingual conversations. Further, we also introduce a task as a proxy for generating a response in a cross-lingual setup. Using parallel data for intermediate fine-tuning also becomes an important addition in the intermediate fine-tuning literature which has largely focused on related monolingual tasks. Our best method leads to an impressive performance on the standard benchmark of the Multilingual WoZ 2.0 dataset \citep{mrksic-etal-2017-semantic} and the recently released parallel MultiWoZ 2.1 dataset \citep{DBLP:journals/corr/abs-2011-06486}. It uses dialogue history and parallel conversational context confirming that our design principles based on conversation history and cross-lingual conversations are important. 
Our methods use 200K parallel movie subtitles \citep{Lison2016OpenSubtitles2016EL} for intermediate training and this data is already available for 1782 language pairs allowing extension to new language pairs. \footnote{Our code is available at \url{https://github.com/nikitacs16/xlift_dst}
}

Our contributions can be summarized as follows: \\
1. To the best of our knowledge, this is the first work to use parallel data for intermediate fine-tuning of multilingual models for multilingual dialogue tasks. We provide strong empirical evidence on four language directions in two datasets for low-resource and zero-shot data scenarios.\\
2. Our proposed intermediate fine-tuning techniques produce data-efficient target language dialogue state trackers. We achieve state-of-the-art results for the zero-shot Multilingual WoZ dataset for most of the metrics and obtain > 20\% improvement on joint goal accuracy with limited labelled data in the target language for the MultiWoZ dataset over the baseline. \\
3. We propose two new intermediate tasks: Cross-lingual dialogue modelling (XDM) and Response masking (RM) that can be extended to other cross-lingual dialogue tasks.

\section{Related Work}
\label{sec:background}

\noindent \textbf{Intermediate fine-tuning of large language models}:
Training deep neural networks on large unlabelled text data to learn meaningful representations has shown remarkable success on several downstream tasks. These representations can be monolingual \citep{qiu2020pretrained} or multilingual \citep{devlin-etal-2019-bert, DBLP:conf/nips/ConneauL19,DBLP:journals/tacl/ArtetxeS19} depending on the underlying training data. These representations are further refined to suit the downstream task by fine-tuning the pretrained model on related data and/or tasks. This ``intermediate'' fine-tuning is done before fine-tuning the task-specific architecture on the downstream task.

In adaptive intermediate fine-tuning, a pretrained model is fine-tuned with the same objectives used during pretraining on data that is closer to the distribution of the target task. This is referred to as task adaptive pretraining (\textbf{TAPT}) if the unlabeled text of the task dataset is used \citep{gururangan-etal-2020-dont, howard-ruder-2018-universal, mehri-etal-2019-pretraining}  and domain adaptive pretraining if unlabelled data of target domain is used  \citep{gururangan-etal-2020-dont, han-eisenstein-2019-unsupervised}. Closer to our problem, \citet{lin2021empirical} also use TAPT for generative dialogue state tracking.  
Another popular method is intermediate task training. Instead of fine-tuning with the objectives used during pretraining of the model, the pretrained model is fine-tuned with single or multiple related tasks as an intermediate step \citep{pruksachatkun-etal-2020-intermediate, phang2019sentence, glavas-vulic-2021-supervised}.  
We refer to the umbrella term of intermediate fine-tuning while discussing our methods.  

Our work uses OpenSubtitles \citep{Lison2016OpenSubtitles2016EL}, a parallel movie subtitle corpus,  as the unlabelled target domain resource. Instead of using the pretrained objectives of the underlying language model directly, we experiment with existing and new objectives to leverage the conversational and cross-lingual nature of the parallel data. 
As there is a dearth of availability of training data for dialogue tasks across different languages, instead of relying on the related task datasets to perform intermediate fine-tuning, we leverage the dialogue data available through OpenSubtitles (See Table \ref{tab:example}). 

\noindent \textbf{Cross-lingual dialogue state tracking}: Dialogue state tracking (DST) is one of the most studied problems in task-oriented conversational systems \citep{mrksic-etal-2017-neural, ren-etal-2018-towards, Chen2018-yy}. The goal of the dialogue state tracker is to accurately identify the user's goals and requests at each turn of the dialogue. These goals and requests are stored in a dialogue state which is predefined based on the ontology of the given domain. For example, the restaurant reservation domain will consist of slot-names like ``price-range'' and values like ``cheap''. Dialogue state tracking has been explored extensively for the monolingual setup but there are limited works for a multilingual setting.

A popular benchmark for cross-lingual dialogue state tracking is the Multilingual WoZ 2.0 dataset \citep{mrksic-etal-2017-semantic} where a dialogue state tracker is trained only on English data and it is evaluated directly for German and Italian dialogue state tracking. \textbf{XL-NBT} \citep{Chen2018-yy}, the first neural cross-lingual dialogue state tracker uses a teacher-student network where the teacher network has access to task labelled data in the source language. The teacher also has access to parallel data which allows it to transfer knowledge to the student network trained in the target language. A couple of recent works resort to code-mixed data augmentation to enhance transfer learning. In Attention-Informed Mixed Language Training \textbf{(AMLT)} \citep{DBLP:conf/aaai/LiuWLXF20}, initially, a dialogue state tracker \citep{mrksic-etal-2017-neural} is trained with English state tracking data. The new code-mixed training data is obtained by replacing the words which receive the highest attention in the given utterance during training of the model with the source language with their respective synonyms in the target language. Another method dubbed as Cross-Lingual Code Switched Augmentation (\textbf{CLCSA}) \citep{DBLP:conf/ijcai/QinN0C20} focuses on the dynamic replacement of source language words with target language words during training.  In this method, the  sentences within a batch are chosen randomly, and then words within these sentences are chosen randomly which are replaced with the synonyms from their target language. This method is state-of-the-art for the Multilingual WoZ dataset. 

Another recent benchmark is the parallel MultiWoZ 2.1 dataset released as a part of the Ninth Dialogue Systems and Technologies Challenge (DSTC-9) \citep{DBLP:journals/corr/abs-2011-06486}. Both the ontology of the dialogue states and the dialogues were translated from English to Chinese using Google Translate and then corrected manually by expert annotators. Similarly, CrossWoZ \citep{DBLP:journals/tacl/ZhuHZZH20}, a Chinese dialogue state tracking dataset was translated into English. The challenge was designed to treat the source dataset as a resource-rich dataset and build a cross-lingual dialogue state tracker which would be evaluated for the low resource target dataset. Instead, all the submissions in the shared task used the translated version of the dataset and treated the problem as a monolingual dialogue state tracking setup. 

We use the Multilingual WoZ dataset and the parallel MultiWoZ dataset to demonstrate the effectiveness of our methods. As there are no existing benchmarks for cross-lingual dialogue state tracking for the parallel MultiWoZ dataset, we use the 
slot-utterance matching belief tracker (SUMBT) \citep{lee-etal-2019-sumbt} as our baseline, which was the state-of-the-art for the English MultiWoZ 2.1 dataset \citep{eric-etal-2020-multiwoz}. The SUMBT model
uses BERT encoder to obtain contextual semantic vectors for the utterances, slot-names, and slot values. It then uses a multi-head attention network to learn the relationship between slot-names and slot-values appearing in the text to predict the dialogue states. 

\section{Intermediate fine-tuning for dialogue tasks}
In this section, we will provide details about the training data used for different intermediate tasks, explain existing and proposed intermediate tasks, and detail their integration into the end task.

\subsection{Adaptive data extraction}
The pretrained language models are often trained on news text or Wikipedia which is different from human conversations \citep{wolf2019transfertransfo}. We choose OpenSubtitles corpus \citep{Lison2016OpenSubtitles2016EL} as the characteristics of this corpus are suitable for our end task.The corpus is huge (beyond 3.2G sentences) and contains parallel movie dialogue data across different language pairs, allowing us to design cross-lingual tasks as well. 
We extract 200K parallel subtitles for every language pair. These are extracted without modifying the sequence of their occurrence in a particular film, as we intend to work on conversations and not sentences in isolation. 

\subsection{Tasks for intermediate fine-tuning}
\input{ft-example}

After extracting the task-related data, we experiment with existing and new intermediate tasks to continue fine-tuning the underlying multilingual representation for the dialogue tasks. These tasks are variants of the \textit{Cloze} task \citep{doi:10.1177/107769905303000401}, where missing words are predicted for a given sentence/context. This task is also known as Masked Language Modelling (MLM) \citep{devlin-etal-2019-bert}. We introduce extensions to the masked language modelling which are more suitable for the dialogue task.  Our task designs are based on (i) interaction between the source and target languages and (ii) interaction between the dialogue history and response. In the rest of the work, the use of the word ``context'' focuses on the role of dialogue history. 

\noindent \textbf{Monolingual dialogue modelling (MonoDM)}:
Dialogue history is an important component of any dialogue task. We select K continuous subtitles from the monolingual subtitles data where K is chosen randomly between 2 to 15 for every example. By choosing a random K, we ensure that the examples contain varied length dialogues as will be the case for any dialogue related task. These examples are created for both the source and the target language and 15\% of the words in each example are masked. 

We now look at cross-lingual intermediate tasks that leverage the parallel data in OpenSubtitles. The following tasks are designed to exploit the contextual information from the dialogue history as well as cross-lingual information through the parallel data. Please see Table \ref{tab:example} for examples.
\\
\noindent \textbf{Translation language modelling (TLM)}:
Translation language modelling (TLM) was introduced while designing the Cross-Lingual Language Model (XLM) \citep{DBLP:conf/nips/ConneauL19}. In TLM, parallel sentences are concatenated and words are masked across them. 
We further explore the importance of longer context in modelling cross-lingual embedding spaces for the conversational setting by concatenating parallel dialogues with K utterances and then masking words randomly on this concatenated text. The hypothesis is that by predicting masked words in different languages simultaneously, the model improves the alignment in its cross-lingual representation space. For the example in Table \ref{tab:example}, the model may learn to align ``bat'' with ``Fledermaus''.
\\
\noindent \textbf{Cross-lingual dialogue modelling (XDM)}: This task focuses on improving cross-lingual context-response representation space.  In TLM, it is difficult to identify if the predicted word used its monolingual context or the bilingual dialogue history. To encourage a cross-lingual interaction between the dialogue history and the response, we concatenate a conversation context (K utterances) from one language and then append the reply to that conversation in the second language. The words are then randomly masked across this chat.

\noindent \textbf{Response masking (RM)}:
We also experiment with a setup that acts as a proxy for generating a response in a cross-lingual setting. The context of the conversation is provided in one language and the task is to predict the words in the response independently in another language. This is a harder task than predicting randomly masked words. 

Both XDM and RM are new designs for intermediate tasks, tailored for cross-lingual dialogue tasks.
We also experimented with combining monolingual and cross-lingual objectives but our pilot experiments did not show any considerable improvement over the individual objectives. For tasks where combining multiple objectives has worked, those tasks required higher reasoning and inference capabilities like coreference resolution or question answering \citep{pruksachatkun-etal-2020-intermediate, aghajanyan2021muppet}.
Such highly specific task data is not available for all languages and even further limited for conversational tasks. We will explore this direction in future. Similarly, our initial experiments suggested that simply combining data from multiple languages for a multilingual intermediate task has lower performance than individual cross-lingual intermediate tasks. Thus, designing multilingual intermediate tasks is far from trivial and we will also explore this in future.

\subsection{Using intermediate fine-tuning for dialogue state tracking}
We create 100K examples for all of the above intermediate tasks for respective language pairs. We use the mBERT \citep{devlin-etal-2019-bert} model as our starting point and continue training the mBERT model with the above tasks separately. 
%Once we have these enriched mBERT models available, these new mBERT models are used as an encoder in the respective dialogue state tracking tasks. 
Thus, all of our reported experiments follow a two-step pipeline procedure where (i) mBERT is fine-tuned with one of the tasks listed as above and then (ii) a dialogue state tracking model, that uses the new mBERT model, is trained with source language training data with or without additional training data of the target language. Finally, the trained dialogue state tracking model is evaluated on the target language. Please see Fig.  \ref{fig:pipeline} for an illustration.

\section{Experiments}

We experiment with the recently released parallel MultiWoZ dataset \citep{DBLP:journals/corr/abs-2011-06486} and the Multilingual WoZ dataset \citep{mrksic-etal-2017-semantic}. As the datasets vary in difficulty and languages, we choose a different amount of target training data and dialogue state tracking architectures for both of them. We briefly provide their description and discuss the results obtained with our methods. 

\subsection{Task description}
\input{tables/en_zh_results}

\noindent \textbf{Parallel MultiWoZ dataset}:
The source dataset MultiWoZ 2.1 \citep{eric-etal-2020-multiwoz} (hence referred as MultiWoZ) is a multi-domain (seven domains) dialogue dataset containing 10K dialogues in English. Both the ontology of the dialogue states and the dialogues were translated from English to Chinese using Google Translate and then corrected manually by expert annotators. Please refer to \citet{DBLP:journals/corr/abs-2011-06486} for further details on dataset creation.  The state language is constant while the conversation language can vary during training and evaluation. This is a more realistic setup as dialogue state can be considered as an intermediate meaning representation which can be language agnostic like SQL. We also use 10\% of the target language training data as part of the training data. As this dataset was recently introduced, there are no models evaluated on the cross-lingual dialogue state tracking setup. Hence, we use the SUMBT architecture \citep{lee-etal-2019-sumbt} trained with vanilla multilingual BERT as our baseline. %Please see the paper for details. 

\noindent \textbf{Multilingual WoZ dataset}: The source dataset WoZ 2.0 \citep{wen-etal-2017-network} is a restaurant reservation dataset in English. The ontology consists of three ``informable'' slots used to inform the system about the user's constraints while looking for a restaurant and seven ``requestable'' slots used to request additional information about a chosen restaurant.
The task is to learn a dialogue state tracker in English and evaluate it directly for German and Italian dialogue state tracking (zero-shot). Unlike the previous setup, we retain the dialogue states in the same language - German utterances will have German dialogue states, to compare with other approaches in the literature. 

We use the state tracker in \citet{DBLP:conf/ijcai/QinN0C20} that treats the problem as a collection of binary prediction tasks, one task for each slot-value combination. The current utterance and the previous dialogue act  are concatenated together and passed through the pretrained multilingual encoder. All the slot value pairs are passed through the encoder to obtain their representations respectively. These representations are then fed into a classification layer. We do not use SUMBT for this dataset as the cross-lingual state tracking performance was not as competitive as other models in the literature. The training details are listed in Appendix A.

\subsection{Metrics}
The metrics used for dialogue state tracking tasks are turn-level and generally include Slot Accuracy, Slot F1, and Joint Goal Accuracy (JGA). Their descriptions are as follows:

\noindent \textbf{Slot Accuracy}: Proportion of the correct slots predicted across all utterances. \\
\noindent \textbf{Slot F1}: Macro-average of F1 score computed over the individual slot-types and slot-values for every turn. \\
\noindent \textbf{Joint Goal Accuracy}: Proportion of examples (dialogue turns) where the predicted dialogue state matches exactly the ground truth dialogue state.

We report Slot F1 and Joint Goal Accuracy for the parallel MultiWoZ dataset. The En state has 135 slot types while the average number of slot types per utterance is 5. When slot accuracy is computed, it also marks all those slots which were not predicted. Consider 130 not predicted slots, 3 correct slots and 2 incorrect slots. By the definition of accuracy, it would be computed as 133/135 = 0.98 which overlooks the two incorrect slots. Thus, we do not report slot accuracy as it is the least indicator of improvement.

We report Joint Goal Accuracy for Multilingual WoZ dataset, where the state only consists of informable slots. Similarly, Slot Accuracy for informable slots and Request Accuracy for requestable slots are also reported, in line with the literature for this task. 

\subsection{Results}

We report the results of models with and without intermediate task learning for the parallel MultiWoZ dataset in Table \ref{tab:mwoz_main} and the Multilingual WoZ dataset in Table \ref{tab:en_de_it}. We compare the performances of our intermediate fine-tuning methods with task-adaptive pretraining (TAPT) to distinguish the design of our intermediate tasks against simply using the task training data. We also compare our methods on Multilingual WoZ with XL-NBT \citep{Chen2018-yy}, Attention Informed Mixed Language Training \citep{DBLP:conf/aaai/LiuWLXF20} and CLCSA \citep{DBLP:conf/ijcai/QinN0C20}.

Our results show that the use of intermediate fine-tuning of a language model is indeed helpful for dialogue state tracking. Further, the use of cross-lingual objectives (XDM, RM, TLM) is indeed superior to task adaptive pretraining (TAPT) and competitive to the monolingual objective (MonoDM) with TLM consistently performing better than all the cross-lingual objective functions in the target language state tracking. This also suggests that the use of bilingual dialogue history (TLM) is superior to the use of cross-lingual context (XDM) or a harder response generation task (RM) for these datasets. 
\input{tables/en_de_it}

In Table \ref{tab:mwoz_main}, we find that even the weakest intermediate fine-tuning setup has 15.3\% and 16.2\% improvement over the vanilla baseline on joint goal accuracy for target languages Zh and En respectively. The best intermediate task (TLM) has an improvement of 20.4\%  and 24.3\% on joint goal accuracy respectively for En $\rightarrow$ Zh and Zh $\rightarrow$ En. The Slot F1 score has similar trends as the joint goal accuracy. Intermediate fine-tuning helps to improve the performance for source language state tracking as well, with monolingual objectives (TAPT, MonoDM) exhibiting a superior performance as they are trained with monolingual task data.  

\noindent\textbf{Comparison with machine translation}: 
As there are no other baselines available for MultiWoZ, we also compare our approach to translation based methods in Table \ref{tab:mwoz_main}. We follow the setup for In-language training, Translate-train, and Translate-test as described in \citet{hu2020xtreme}. In In-language training, we fine-tune the mBERT model directly with target language training data. For the Translate-train models, we first translate the source language training data of the dialogue task into the target language and then train a dialogue state tracking model with mBERT on the translated target language data. In Translate train, the dialogue state tracking model is trained with the source language data on source language BERT. At test time, the target language instances are translated into the source language to predict the dialogue states for these given instances. Our machine translation models are large transformer models \citep{Vaswani2017-sz} trained on Paracrawl data \citep{banon-etal-2020-paracrawl} for En $\rightarrow$ Zh and Zh $\rightarrow$ En respectively. Our setup improves over the Translate-test approach which uses these additional translation models and monolingual BERT models. We also find that Translate Train and In-language training find this setup difficult as the model would map a target language utterance to a source language state instead of a target language state. Further, following guidelines from \citet{hu2020xtreme}, these models are trained with multilingual BERT which is trained on 108 languages, leading to a noisier representation space than a monolingual BERT. Overall, we find that the scores are higher for Zh $\rightarrow$ En than En $\rightarrow$ Zh. We speculate this trend is due to the presence of translationese when using Zh as the source language as the dataset is originally in English then translated to Chinese, in line with the observations from neural machine translation literature \citep{edunov-etal-2020-evaluation}. 

\noindent\textbf{Additive effect of TLM with CLCSA}: In Table \ref{tab:en_de_it}, we find that TLM has 27.5\% and 24.3\% improvement over the vanilla baseline on joint goal accuracy for De and It respectively. It also has superior performances over baselines from the literature except for the CLCSA method. The CLCSA method uses dynamic code-mixed data for training the state tracker. We observe that using TLM with the CLCSA model has an additive effect, providing an improvement over a model which does not use the model with TLM as an intermediate fine-tuning task. Please note that our experiments for both CLCSA and CLCSA + TLM used an uncased version of multilingual BERT as opposed to the cased version of multilingual BERT in the original CLCSA results as it has better performance. We also find that RM is not best suited for this task suggesting that response prediction is not a suitable intermediate task for simple scenarios of the WoZ dataset. 

\section{Analysis}
We analyse the outputs from the state tracker and design choices for the intermediate tasks. We also provide insights into the difficulty of conducting zero-shot transfer learning using the SUMBT architecture for the MultiWoZ dataset. 
\subsection{Qualitative analysis}
We manually analyzed the predicted dialogue states for 200 chats from these models for the MultiWoZ dataset. Overall, we found that models trained with intermediate tasks improve over the vanilla baselines in detecting cuisine names, names of restaurants, and time periods for booking (taxi/restaurant). All models show some confusion in detecting whether a location corresponds to arrival or departure. We observe that predicting a dialogue state wrong at an earlier stage has a cascading effect of errors on the later dialogue states. For the Multilingual WoZ dataset, the baseline models struggled to identify less frequent cuisines. There was confusion between predicting ``cheap'' and ``moderate'' in the target languages. These errors were reduced with intermediate fine-tuning. Please see examples in Appendix C.

\subsection{Investigating zero-shot transfer for MultiWoZ dataset}
\label{sec:analysis_zero_shot}
We make a case for using 10\% of training data in the target language and retaining the language of the source state for the MultiWoZ dataset. We illustrate different training data choices in Table \ref{tab:amount_of_data}. We currently look at the En $\rightarrow$ Zh setup. 
\input{tables/amount_of_data}

The zero-shot setup is difficult for the models - with the vanilla baseline model, it seems nearly impossible to learn a dialogue state tracker for Chinese. 
Even with TLM, while there is an improvement in the multilingual representation space, it is not adequate for a generalized transfer across languages. However, when a pretrained model which is fine-tuned with a cross-lingual objective, is trained with as little as 1\% labelled target language training data (84 chats), we observe 19.3\% improvement over the joint goal accuracy for the target language over the zero-shot vanilla baseline. This also indicates the data efficiency of the cross-lingual intermediate fine-tuning. With the increase in target training data, the performance for the target language also improves while degrading the source language performance. 

We also found that using the target language states during evaluation has lower performance than source language dialogue states for this dataset while using the SUMBT model. Using a dialogue state tracker trained with TLM on zero-shot setup had joint goal accuracy of 1\%. We recommend mapping the dialogue states from the source language to the target language directly for use cases that require the dialogue state to be predicted in the target language.

\subsection{Analysis of intermediate tasks}
We analyse the design choices for the intermediate tasks - domain and amount of intermediate training data and use of dialogue history. 
\subsubsection{Domain of adaptive task data}

We considered the parallel document level data released for the WMT'19 challenge \citep{ws-2019-machine}. We look at the En-Zh parallel data consisting of news articles that are aligned by paragraphs. 
\input{tables/news_chat}

We fine-tune the mBERT model with the TLM task for parallel paragraphs. We report our results for the MultiWoZ dataset in Table \ref{tab:news-chat}.
We find that using dialogue data has a slight advantage over using parallel news text as seen in Table \ref{tab:news-chat}. This suggests that cross-lingual alignment itself is largely responsible for the increase in the joint goal accuracy over the baseline than the domain of the intermediate task data. Nevertheless, we recommend the use of OpenSubtitles for intermediate task data as it not only performs better but also is available for 1782 language pairs. 

\subsubsection{Amount of intermediate task data}
We used a fixed number of examples for the intermediate fine-tuning. We now vary the amount of intermediate task data and study its performance on the downstream task.
\input{tables/amount_of_interm_data}
As seen from Table \ref{tab:amount_of_intermediate_data}, our setup that uses examples with 200K data has the best or second-best performance across the target languages. There is indeed an increase in performance for target language En with 800K sentences, but fine-tuning a model with 800K sentences also 4x additional GPU training time. We find that the performance drop in addition or removal of intermediate examples is not extreme. This prompts us to design better cross-lingual objectives that can reduce the intermediate data requirement. 

\subsubsection{Utterance-level v/s dialogue history}
We emphasized using dialogue history nformation while designing intermediate tasks. 
%for cross-lingual transfer on downstream dialogue tasks.
For ablation studies, we fine-tune mBERT with utterance-level intermediate tasks. To replicate the utterance-level version of MonoDM (referred to as MonoDM-chat here), the training data for MonoDM-utterance consists of 100K utterances chosen randomly from the OpenSubtitles data, with equal English and Chinese examples. Similarly, TLM-utterances also uses 100K examples with parallel utterances chosen randomly. 
\input{tables/context_utterances}
The results in Table \ref{tab:context_utterances} show that the use of dialogue history is important as both MonoDM-sent and TLM-sent have lower performance than MonoDM-chat and TLM-chat respectively. We observe a similar trend for the Multilingual WoZ dataset (reported in Appendix B).

\section{Conclusion}
We demonstrated the effectiveness of cross-lingual intermediate fine-tuning of pretrained multilingual language models for the task of cross-lingual dialogue state tracking. We experimented with existing intermediate tasks and introduced two new cross-lingual intermediate tasks based on the parallel and dialogue-level nature of the movie subtitles corpus. Our best method had significant improvement in performance for the parallel MultiWoZ dataset and Multilingual WoZ dataset. We also demonstrated the data efficiency of our methods. 

Our intermediate tasks were trained on a generic dataset unlike the related high resource tasks used in \citet{phang-etal-2020-english}. As OpenSubtitles is available for 1782 language pairs, we speculate that using these cross-lingual intermediate tasks will be effective for languages where a collection of large training datasets for dialogue tasks is not feasible. We speculate that this setup can be useful for cross-lingual domain transfer too - when such benchmark becomes available for dialogue tasks. We hope that our method can serve as a strong baseline for future work in multilingual dialogue.

\section*{Acknowledgements}
We would like to thank Liane Guillou for feedback on the experiment setup for this work. We thank Barry Haddow for providing us with the machine translation models. We also thank Laurie Burchell, Agostina Calabrese, Tom Hosking, and the anonymous reviewers for their insightful comments and suggestions. This work was supported in part by the UKRI Centre for Doctoral Training in Natural Language Processing, funded by the UKRI (grant EP/S022481/1) and the University of Edinburgh (Moghe). The authors gratefully acknowledge Huawei for their support (Moghe). 

\bibliography{anthology,custom}
\bibliographystyle{acl_natbib}
\clearpage
\appendix
\section{Reproducibility Details}

\textbf{Hyperparameters}: All the intermediate fine-tuning models were trained with HuggingFace's transformers library \citep{wolf-etal-2020-transformers}. We followed the guidelines from \citet{phang-etal-2020-english} to select the hyperparameters. The fine-tuning was carried out for 20 epochs.  The batch size was between $\{4, 8\}$. The rest configuration was kept as default in the library. \\
For the SUMBT model, the LSTM size was varied between $\{100, 300\}$, the learning rate between $\{1e-4, 1e-5, 5e-5\}$, and batch size between $\{3, 4, 12\}$. Rest hyperparameters were kept as default as the original work.  The final configurations were chosen based on the joint goal accuracy for the development set. The training was carried out for 100 epochs as default with patience of 10 epochs. For the Multilingual WoZ experiments, we followed the hyperparameters listed in \citet{DBLP:conf/ijcai/QinN0C20}
\\
All of our hyperparameters for all the experiments will be made available as config files. We use code from \citet{zhu-etal-2020-convlab} for the SUMBT model and \citet{DBLP:conf/ijcai/QinN0C20} for the CLCSA model.

\textbf{Training details}: Intermediate fine-tuning takes approx 14 hours on RTX 2080 Ti, training a SUMBT model takes approx six hours, and the base architecture for Multilingual WoZ takes around three hours. The training hours on a different GPU may vary. The inference time for the SUMBT model on the MultiWoZ dataset is 4 minutes while that of the Multilingual WoZ is a minute per language. Similarly, the GPU memory for intermediate fine-tuning and SUMBT takes up the entire ram of RTX 2080 Ti ( approx 11 GB) and the Multilingual WoZ experiments occupy 7 GB RAM. All the experiments require a single GPU. The parameters in the mBERT model are approx 178M. The parameters in the dialogue state trackers without the mBERT model are approx 5.2 M and 0.1 M for the MultiWoZ dataset and Multilingual WoZ dataset respectively.

\textbf{{Dataset details}}:
The dialogue state tracking datasets are available at the code repositories of \citet{zhu-etal-2020-convlab} and \citet{DBLP:conf/ijcai/QinN0C20} respectively.
The OpenSubtitles corpus can be obtained from the corpus website \footnote{ \url{https://opus.nlpl.eu/OpenSubtitles-v2018.php}} which is based on the subtitles website\footnote{\url{http://www.opensubtitles.org/}}. 
We will release the extracted examples and their variants as well. 
Please see Table \ref{tab:data_stats} for statistics. While creating the 10\% of the labelled target language data, all the domains the in the MultiWoZ data were included according to their proportion in the original training data. 

\begin{table}[]
\small
\centering
%\resizebox{\linewidth}{!}{
\begin{tabular}{|l|r|r|}
\hline
Stat                 & MultiWoZ & Multilingual WoZ \\ \hline
\#Train Chats        & 8434     & 600              \\ \hline
\#Dev Chats          & 1000     & 200              \\ \hline
\#Test Chats         & 1000     & 400              \\ \hline
Train       & En, Zh   & En               \\ \hline
Evaluation  & Zh, En   & De, It           \\ \hline
\end{tabular}
\caption{Datasets Statistics}
\label{tab:data_stats}
\end{table}

\section{Utterance v/s Dialogue history for Multilingual WoZ}

We report the importance of using dialogue history in Table \ref{tab:woz_utt}.

\begin{table}[]
\small
\centering
\resizebox{\linewidth}{!}{
\begin{tabular}{|c|c|c|c|c|c|c|}
\hline
\multirow{2}{*}{\begin{tabular}[c]{@{}c@{}}Intermediate\\ Fine-tuning\end{tabular}} & \multicolumn{3}{c|}{De} & \multicolumn{3}{c|}{It} \\ \cline{2-7} 
 &
  \begin{tabular}[c]{@{}c@{}}Slot \\ Acc\end{tabular} &
  \begin{tabular}[c]{@{}c@{}}Joint \\ Acc\end{tabular} &
  \begin{tabular}[c]{@{}c@{}}Request \\ Acc\end{tabular} &
  \begin{tabular}[c]{@{}c@{}}Slot\\  Acc\end{tabular} &
  \begin{tabular}[c]{@{}c@{}}Joint \\ Acc\end{tabular} &
  \begin{tabular}[c]{@{}c@{}}Request \\ Acc\end{tabular} \\ \hline
none                                                                                & 57.6   & 15     & 75.3  & 54.6   & 12.6   & 77.3  \\ \hline
MonoDM-sent                                                                           & 59.2   & 7.5    & 88    & 57     & 2.43   & 83    \\ \hline
MonoDM-chat                                                                           & 83.4   & 14.4   & 90.3  & 63.6   & 14.1   & 90.2  \\ \hline
TLM-sent                                                                            & 73     & 33.2   & 91.4  & 60.3   & 8.7    & 89    \\ \hline
TLM-chat                                                                            & 75.6   & 42.5   & 90.2  & 72.3   & 36.9   & 90    \\ \hline
\end{tabular}}
\caption{Comparison of amount of dialogue history used in intermediate tasks and evaluated for target languages in Multilingual WoZ. Sent - sentences. Use of chats in intermediate fine-tuning tasks is beneficial. }
\label{tab:woz_utt}
\end{table}
\section{Qualitative Examples}
\input{tables/example_output}
In Table \ref{tab:example_outputs}, the first example demonstrates how TLM can identify named entities such as names of restaurants that the baseline could not predict. Similarly, the baseline has a higher error rate detecting the dialogue states with numbers, as seen in examples one and two. The third example is a continuation of the conversation in the second example. Note that the baseline model is now capable of predicting all the new dialogue states in this example. But it is penalized as it could not predict the train-arriveby state at the start of the conversation leading to cascading of errors.

\end{document}

%% file: ft-example.tex
\begin{table*}[]
\small
\resizebox{\textwidth}{!}{
\begin{tabular}{|l|l|l|l|l|l|}
\hline
\multicolumn{3}{|l|}{Subtitle (En)}    & \multicolumn{3}{l|}{Subtitle (De)}    \\ \hline
\multicolumn{3}{|p{3.1 in}|}{\begin{tabular}[c]{@{}l@{}}Who is it, Martin? A bat, Professor. Very big and black. Don't \\ waste your pellets. It's no use. \textit{You'll never harm that bat.} \\\end{tabular}} &
  \multicolumn{3}{l|}{\begin{tabular}[c]{@{}l@{}}Wer ist denn da, Martin? Eine Fledermaus, Herr Professor.  Sehr \\ gro{\ss} und pechschwarz. Verschwenden Sie kein Schrot darauf. Es ist \\ zwecklos. \textit{Dieser Fledermaus k{\"o}nnen Sie nichts anhaben.}\\\end{tabular}} \\ \hline
\multicolumn{2}{|c|}{\textbf{TLM}} & \multicolumn{2}{c|}{\textbf{XDM}} & \multicolumn{2}{c|}{\textbf{RM}} \\ \hline 
\multicolumn{2}{|l|}{\begin{tabular}[c]{@{}l@{}}Who is it, Martin A {[}MASK{]} \dots \\  {[}MASK{]} that bat. {[}MASK{]} ist denn da,\\  Martin? \dots \textit{k{\"o}nnen Sie nichts} {[}MASK{]}.\end{tabular}} &
  \multicolumn{2}{l|}{\begin{tabular}[c]{@{}l@{}}Who is it, Martin? A {[}MASK{]} \dots \\ of no use. \textit{Dieser Fledermaus k{\"o}nnen} \\ \textit{Sie nichts} {[}MASK{]}.\end{tabular}} &
  \multicolumn{2}{l|}{\begin{tabular}[c]{@{}l@{}}Who is it, Martin? A bat, Professor  \dots \\ It's no use. {[}MASK{]}{[}MASK{]}{[}MASK{]}\\ {[}MASK{]} {[}MASK{]} {[}MASK\}\end{tabular}} \\ \hline
\end{tabular}
}
\caption{Examples for different cross-lingual intermediate tasks. The top row contains the parallel text converted into examples. The intermediate task is to predict the {[}MASK{]} words. TLM - Translation Language Modelling, XDM - Cross-lingual Dialogue Modelling, RM - Response Masking. \textit{Italics} is the response in the given chat.}
\label{tab:example}
\end{table*}

%% file: tables/en_zh_results.tex
% \usepackage{multirow}
\begin{table*}[h]
\small
\centering
\begin{tabular}{|c|c|c|c|c||c|c|c|c|c|c|}
\hline
 \begin{tabular}[c]{@{}l@{}}Intermediate \\ Fine-tuning\end{tabular} & \multicolumn{2}{c|}{\begin{tabular}[c]{@{}c@{}}Source Language\\ En\end{tabular}} & \multicolumn{2}{c|}{\begin{tabular}[c]{@{}c@{}}Target Language\\ Zh\end{tabular}} & \multicolumn{2}{c|}{\begin{tabular}[c]{@{}c@{}}Source Language\\ Zh\end{tabular}} &
 \multicolumn{2}{c|}{\begin{tabular}[c]{@{}c@{}}Target Language\\ En \end{tabular}} &
 \multicolumn{2}{c|}{\begin{tabular}[c]{@{}c@{}}Target Language\\ Avg Gain\end{tabular}} \\ \hline
          & JGA        & Slot F1        & JGA       & Slot F1        & JGA        & Slot F1        & JGA        & Slot F1       & JGA        & Slot F1 \\ \hline
none             & 15.5          & 66.4          & 12.3         & 60.9          & 17.1        & 73.4        & 16.8          & 73.2      & 00.0 & 00.0    \\ \hline
  TAPT             & \textbf{44.3} & \textbf{88.6} & 27.6          & 78.7          & 40.0          & 84.8        & 33.0            & 81.3  & 15.7 & 12.9        \\ \hline
  MonoDM             & 39.0            & 85.6          & 28.2          & 78.8          & \textbf{44.0} & \textbf{88.0} & \textbf{41.7} & 87.3  & 20.4 & 16.0        \\ 
 \hline
  XDM              & 41.7          & 87.3          & 29.6          & 80.3          & 43.6        & 88.1        & 39.3          & 86.2   &19.9 & 16.2       \\ \hline
  RM & 42.5          & 88.0            & 32.5          & 82.0            & 42.0          & 87.5        & 38.4          & 85.4          & 20.9 & 16.8 \\ \hline
  TLM         & 44            & 88.4          & \textbf{32.7} & \textbf{82.4} & 42.7        & 88.7        & 41.1          & \textbf{87.7} & \textbf{22.3} & \textbf{18.0}\\\hline

\hline
In-language training          & -  & -            & 15.8  & 70.2                & -  & -                & 17.9  & 74.5  & 01.2 & 03.7              \\ \hline

Translate-Train               & - & -             & 11.1 & 54.2                & - & -                & 15.8 & 71.4     & -1.1 & -4.2           \\ \hline
Translate-Test                & - & -             & 26.5 & 77.0                & -  & -                & 22.7 & 77.4    & 10.0 & 10.1            \\ \hline
 
\end{tabular}
\caption{Performance on the parallel MultiWoZ dataset using encoders with  various intermediate fine-tuning strategies and trained with 100\% source and 10\% target language dialogue state tracking data. \textbf{Bold} marks the best within each column. JGA - Joint goal accuracy. The last two columns indicate average gain over mBERT-none for target languages.}
\label{tab:mwoz_main}
\end{table*}

%% file: tables/en_de_it.tex
% \usepackage{multirow}
\begin{table*}
\small
\centering
\begin{tabular}{|l|l|l|l|l|l|l|l|l|}
\hline
\begin{tabular}[c]{@{}l@{}}Multilingual Model/ \\ Method \\ \end{tabular} &
  \begin{tabular}[c]{@{}l@{}}Intermediate \\ Task Training\end{tabular} &
  \multicolumn{3}{c|}{\begin{tabular}[c]{@{}c@{}}Target Language\\ De\end{tabular}} &
  \multicolumn{3}{c|}{\begin{tabular}[c]{@{}c@{}}Target Language\\ It\end{tabular}} & \begin{tabular}[c]{@{}l@{}}Average \\ Gain \\ \end{tabular}\\ \hline
 &
   &
  \begin{tabular}[c]{@{}l@{}}Slot \\ Acc\end{tabular} &
  \begin{tabular}[c]{@{}l@{}}Joint \\ Acc\end{tabular} &
  \begin{tabular}[c]{@{}l@{}}Request\\ Acc\end{tabular} &
  \begin{tabular}[c]{@{}l@{}}Slot\\ Acc\end{tabular} &
  \begin{tabular}[c]{@{}l@{}}Joint \\ Acc\end{tabular} &
  \begin{tabular}[c]{@{}l@{}}Request\\ Acc\end{tabular} & \begin{tabular}[c]{@{}l@{}}Joint\\ Acc\end{tabular} \\ \hline
    XL-NBT \citep{Chen2018-yy}           & N/A         & 55   & 30.8 & 68.4 & 72   & 41.2 & 81.2 & 22.2 \\ 
    AMLT \citep{DBLP:conf/aaai/LiuWLXF20}    & N/A              & 70.7 & 34.3 & 87   & 71.4 & 33.3 & 84.9 & 20.0 \\ \hline
    \multirow{6}{*}{mBERT} & none        & 57.6 & 15   & 75.3 & 54.6 & 12.6 & 77.3 & 00.0 \\ 
                      & TAPT              & 68.4 & 24.8 & 89   & 67.5 & 22.6 & 83.8 & 09.9 \\ \cline{2-9}
                      & MonoDM             & 83.4 & 14.4 & 90.3 & 63.6 & 14.1 & 90.2 & 00.4\\ 
                      & XDM              & 69.7 & 27.5 & 90   & 68   & 21.5 & 89.1 & 10.7 \\ 
                      & RM &  58    & 8.6     & 81.6      & 61.6     & 11.3     & 76.4    & -3.8 \\ 
                      & TLM         & 75.6 & 42.5 & 90.2 & 72.3 & 36.9 & 90  & 25.9 \\ \hline
\multirow{2}{*}{CLCSA \citep{DBLP:conf/ijcai/QinN0C20}} & none             & 83.2   & 62.6 & \textbf{96.1}   & 84 &\textbf{67.6} & \textbf{95.5} & 51.3 \\ 
                      & TLM         & \textbf{85.2} & \textbf{65.8} & 94.4 & \textbf{84.3} & 66.9 & \textbf{95.5} & \textbf{52.5}\\\hline

\end{tabular}
\caption{Zero-shot results of the target languages of Multilingual WoZ 2.0 dataset with and without using various intermediate fine-tuning strategies when trained with English task data. Acc - Accuracy. The last column is average gain over joint accuracy for both the languages over the mBERT-none model. Please see text for details of the methods. \textbf{Bold} indicates the best score in that column. Intermediate fine-tuning is also useful for zero-shot transfer and cross-lingual intermediate fine-tuning (TLM) has the best performance.}
\label{tab:en_de_it}
\end{table*}

%% file: tables/amount_of_data.tex
% \usepackage{multirow}
\begin{table}[h]
\small
\centering
\resizebox{\linewidth}{!}{
\begin{tabular}{|c|c|c|c|c|c|}
\hline
\multirow{2}{*}{\begin{tabular}[c]{@{}c@{}}Intermediate\\ Fine-tuning\end{tabular}} &
  \multirow{2}{*}{\begin{tabular}[c]{@{}c@{}}Target \\ Data (\%) \end{tabular}} &
  \multicolumn{2}{c|}{\begin{tabular}[c]{@{}c@{}}Source \\ En\end{tabular}} &
  \multicolumn{2}{c|}{\begin{tabular}[c]{@{}c@{}}Target\\ Zh\end{tabular}} \\ \cline{3-6} 
     &    & JGA  & Slot F1 & JGA  & Slot F1 \\ \hline
none & 0  & 16.8          & 73.4          & 01.9  & 14.8 \\ \hline
TLM  & 0  & 43.9          & 88.7          & 05.1  & 40.7 \\ \hline
TLM  & 1  & 45.1          & 89.1          & 21.2  & 71.9 \\ \hline
TLM  & 5  & 44.2          & 88.9          & 31.3  & 82.1 \\ \hline
TLM  & 10 & 44.0          & 88.4          & 32.7  & 82.4 \\ \hline
none & 10 & 15.5          & 66.4          & 12.3 & 60.9 \\ \hline
\end{tabular}
}

\caption{Comparing different proportions of target state tracking data  along with En training data for En $\rightarrow$ Zh MultiWoZ dataset. Zero-shot setup is difficult for this task but it can be improved with limited Zh data and intermediate fine-tuning }
\label{tab:amount_of_data}
\end{table}

%% file: tables/news_chat.tex
\begin{table}[h]
\small
\centering
\resizebox{\linewidth}{!}{
\begin{tabular}{|c|c|c|c|c|c|}
\hline
\multirow{2}{*}{\begin{tabular}[c]{@{}l@{}}Intermediate\\ fine-tuning \end{tabular}} & \multirow{2}{*}{\begin{tabular}[c]{@{}l@{}}Intermediate \\ task data\end{tabular}}  & \multicolumn{2}{c|}{\begin{tabular}[c]{@{}c@{}}Target\\ Zh\end{tabular}} &
  \multicolumn{2}{c|}{\begin{tabular}[c]{@{}c@{}}Target\\ En\end{tabular}} \\ \cline{3-6} 
    &     & JGA   & Slot F1 & JGA  & Slot F1 \\ \hline
none &     -                    & 12.3 & 60.9 & 16.8 & 73.2\\ \hline
TLM  & Movie subtitles     & 32.7 & 82.4 & 41.1 & 87.7\\ \hline
TLM  & News Text           & 32.0 & 81.8 & 41.5 & 87.2 \\ \hline
\end{tabular}}
\caption{Investigating the domain of intermediate task data evaluated on the target languages of parallel MultiWoZ data. Intermediate fine-tuning on movie subtitles is slightly advantageous over news texts }
\label{tab:news-chat}
\end{table}

%% file: tables/amount_of_interm_data.tex
\begin{table}[]
\small
\centering
\begin{tabular}{|c|c|c|c|c|}
\hline
\begin{tabular}[c]{@{}c@{}}Amount of \\ Intermediate \\ Data\end{tabular} & \multicolumn{2}{c|}{\begin{tabular}[c]{@{}c@{}}Target \\ Zh\end{tabular}} & \multicolumn{2}{c|}{\begin{tabular}[c]{@{}c@{}}Target\\ En\end{tabular}} \\ \hline
0.5 x                                                                     & 29.6                                & 80.3                                & 38.3                                & 85.8                               \\ \hline
x                                                                         & 32.7                                & 82.4                                & 41.1                                & 87.7                               \\ \hline
2 x                                                                       & 29.4                                & 80.6                                & 40.8                                & 87.1                               \\ \hline
4 x                                                                       & 29.3                                & 81.2                                & 43.9                                & 88.2                               \\ \hline
\end{tabular}
\caption{Comparison of amount of intermediate task data when used with TLM on MultiWoZ. x: examples created with 200K data. Using 200K data is indeed optimal}
\label{tab:amount_of_intermediate_data}
\end{table}

%% file: tables/context_utterances.tex
% \usepackage{multirow}
\begin{table}[]
\small
\centering
\begin{tabular}{|c|c|c|c|c|}
\hline
\multirow{2}{*}{\begin{tabular}[c]{@{}c@{}}Intermediate\\ Fine-tuning\end{tabular}} &
  \multicolumn{2}{c|}{\begin{tabular}[c]{@{}c@{}}Target\\ Zh\end{tabular}} &
  \multicolumn{2}{c|}{\begin{tabular}[c]{@{}c@{}}Target\\ En\end{tabular}} \\ \cline{2-5} 
         & JGA   & Slot F1 & JGA  & Slot F1 \\ \hline
%none     & 12.33 & 60.9    & 16.8 & 73.2    \\ \hline
MonoDM-sent & 24.2  & 75.4    & 30.9 & 81.6    \\ \hline
MonoDM-chat     & 28.2  & 78.8    & 41.7 & 87.3    \\ \hline
TLM-sent & 31.2  & 81.3    & 34.7 & 83.2    \\ \hline
TLM-chat & 32.7  & 82.4    & 41.1 & 87.7    \\ \hline
\end{tabular}

\caption{Comparison of amount of dialogue history used in intermediate tasks and evaluated for target languages in MultiWoZ. Sent - sentences. Use of chats in intermediate fine-tuning tasks is beneficial. }
\label{tab:context_utterances}
\end{table}

%% file: tables/example_output.tex
\begin{table*}[]
\begin{CJK*}{UTF8}{gbsn}
\small
\centering
\resizebox{\linewidth}{!}{%
\begin{tabular}{|l|l|l|l|l|}
\hline
Setup & Context                                                                                                                                                                                                                                                                                             & None                                                                                                                                                                 & TLM                                                                                                                                                                                         & Ground Truth                                                                                                                                                                                \\ \hline
En-Zh & \begin{tabular}[c]{@{}l@{}}剑桥主轴宾馆是3星级的家庭旅馆。\\ 它在南部地区。您想预定一个房间吗？\\ (The Bridge Guest House is a 3 star guesthouse.\\ It is in the south area.Would you like to book a\\ room ?)\end{tabular}                                                                                                          & \begin{tabular}[c]{@{}l@{}}hotel-stars-4\\ \\ hotel-type-guesthouse\end{tabular}                                                                                     & \begin{tabular}[c]{@{}l@{}}hotel-stars-3\\ hotel-type-guesthouse\\ hotel-name-bridge guest house\\ hotel-area-south\end{tabular}                                                            & \begin{tabular}[c]{@{}l@{}}hotel-stars-3\\ hotel-type-guesthouse\\ hotel-name-bridge guest house\\ hotel-area-south\end{tabular}                                                            \\ \hline
En-Zh & \begin{tabular}[c]{@{}l@{}}我需要从剑桥乘火车，我必须在17：00 /\\ 之前到达目的地\\ (I need to take a train from cambridge,\\ I need to arrive at my destination by 17:00)\end{tabular}                                                                                                                                     & \begin{tabular}[c]{@{}l@{}}train-destination-norwich\\ train-day-saturday\\ train-departure-cambridge\end{tabular}                                          & \begin{tabular}[c]{@{}l@{}}train-destination-norwich\\ train-day-saturday\\ train-arriveby-17:00\\ train-departure-cambridge\end{tabular}                                                   & \begin{tabular}[c]{@{}l@{}}train-destination-norwich\\ train-day-saturday\\ train-arriveby-17:00\\ train-departure-cambridge\end{tabular}                                          \\ \hline
En-Zh & \begin{tabular}[c]{@{}l@{}}您还能找到我一个吃东西的地方吗？\\ 我当然可以！您是否在寻找特定的地区和食物类型？\\ 请给我在中心的印度餐厅。\\ (Can you also find me a place to get some food? \\ I sure can! Do you have a specific area and type of \\ food you are looking for? \\ I would like an indian restaurant in the centre, please)\end{tabular} & \begin{tabular}[c]{@{}l@{}}train-destination-norwich\\ train-day-saturday\\ train-departure-cambridge\\ restaurant-food-indian\\ restaurant-area-centre\end{tabular} & \begin{tabular}[c]{@{}l@{}}train-destination-norwich\\ train-day-saturday\\ train-arriveby-17:00\\ train-departure-cambridge\\ restaurant-food-indian\\ restaurant-area-centre\end{tabular} & \begin{tabular}[c]{@{}l@{}}train-destination-norwich\\ train-day-saturday\\ train-arriveby-17:00\\ train-departure-cambridge\\ restaurant-food-indian\\ restaurant-area-centre\end{tabular} \\ \hline
\end{tabular}}
\end{CJK*}
\caption{Example outputs from the En-Zh systems. We demonstrate how TLM improves in detecting named entities, numbers, and prevents cascading effect of predicting an example wrong at the start of the dialogue.}
\label{tab:example_outputs}
\end{table*}

%% file: emnlp2021.bbl
\begin{thebibliography}{36}
\expandafter\ifx\csname natexlab\endcsname\relax\def\natexlab#1{#1}\fi

\bibitem[{Aghajanyan et~al.(2021)Aghajanyan, Gupta, Shrivastava, Chen,
  Zettlemoyer, and Gupta}]{aghajanyan2021muppet}
Armen Aghajanyan, Anchit Gupta, Akshat Shrivastava, Xilun Chen, Luke
  Zettlemoyer, and Sonal Gupta. 2021.
\newblock \href {http://arxiv.org/abs/2101.11038} {Muppet: Massive multi-task
  representations with pre-finetuning}.

\bibitem[{Artetxe and Schwenk(2019)}]{DBLP:journals/tacl/ArtetxeS19}
Mikel Artetxe and Holger Schwenk. 2019.
\newblock \href {https://transacl.org/ojs/index.php/tacl/article/view/1742}
  {Massively multilingual sentence embeddings for zero-shot cross-lingual
  transfer and beyond}.
\newblock \emph{Trans. Assoc. Comput. Linguistics}, 7:597--610.

\bibitem[{Ba{\~n}{\'o}n et~al.(2020)Ba{\~n}{\'o}n, Chen, Haddow, Heafield,
  Hoang, Espl{\`a}-Gomis, Forcada, Kamran, Kirefu, Koehn, Ortiz~Rojas,
  Pla~Sempere, Ram{\'\i}rez-S{\'a}nchez, Sarr{\'\i}as, Strelec, Thompson,
  Waites, Wiggins, and Zaragoza}]{banon-etal-2020-paracrawl}
Marta Ba{\~n}{\'o}n, Pinzhen Chen, Barry Haddow, Kenneth Heafield, Hieu Hoang,
  Miquel Espl{\`a}-Gomis, Mikel~L. Forcada, Amir Kamran, Faheem Kirefu, Philipp
  Koehn, Sergio Ortiz~Rojas, Leopoldo Pla~Sempere, Gema
  Ram{\'\i}rez-S{\'a}nchez, Elsa Sarr{\'\i}as, Marek Strelec, Brian Thompson,
  William Waites, Dion Wiggins, and Jaume Zaragoza. 2020.
\newblock \href {https://doi.org/10.18653/v1/2020.acl-main.417} {{P}ara{C}rawl:
  Web-scale acquisition of parallel corpora}.
\newblock In \emph{Proceedings of the 58th Annual Meeting of the Association
  for Computational Linguistics}, pages 4555--4567, Online. Association for
  Computational Linguistics.

\bibitem[{Bojar et~al.(2019)Bojar, Chatterjee, Federmann, Fishel, Graham,
  Haddow, Huck, Yepes, Koehn, Martins, Monz, Negri, N{\'e}v{\'e}ol, Neves,
  Post, Turchi, and Verspoor}]{ws-2019-machine}
Ond{\v{r}}ej Bojar, Rajen Chatterjee, Christian Federmann, Mark Fishel, Yvette
  Graham, Barry Haddow, Matthias Huck, Antonio~Jimeno Yepes, Philipp Koehn,
  Andr{\'e} Martins, Christof Monz, Matteo Negri, Aur{\'e}lie N{\'e}v{\'e}ol,
  Mariana Neves, Matt Post, Marco Turchi, and Karin Verspoor, editors. 2019.
\newblock \href {https://www.aclweb.org/anthology/W19-5200} {\emph{Proceedings
  of the Fourth Conference on Machine Translation (Volume 1: Research
  Papers)}}. Association for Computational Linguistics, Florence, Italy.

\bibitem[{Chen et~al.(2018)Chen, Chen, Su, Wang, Yu, Yan, and
  Wang}]{Chen2018-yy}
Wenhu Chen, Jianshu Chen, Yu~Su, Xin Wang, Dong Yu, Xifeng Yan, and
  William~Yang Wang. 2018.
\newblock {XL-NBT}: A cross-lingual neural belief tracking framework.
\newblock In \emph{Proceedings of the 2018 Conference on Empirical Methods in
  Natural Language Processing}, pages 414--424, Stroudsburg, PA, USA.
  Association for Computational Linguistics.

\bibitem[{Conneau and Lample(2019)}]{DBLP:conf/nips/ConneauL19}
Alexis Conneau and Guillaume Lample. 2019.
\newblock \href
  {http://papers.nips.cc/paper/8928-cross-lingual-language-model-pretraining}
  {Cross-lingual language model pretraining}.
\newblock In \emph{Advances in Neural Information Processing Systems 32: Annual
  Conference on Neural Information Processing Systems 2019, NeurIPS 2019, 8-14
  December 2019, Vancouver, BC, Canada}, pages 7057--7067.

\bibitem[{Devlin et~al.(2019)Devlin, Chang, Lee, and
  Toutanova}]{devlin-etal-2019-bert}
Jacob Devlin, Ming-Wei Chang, Kenton Lee, and Kristina Toutanova. 2019.
\newblock \href {https://doi.org/10.18653/v1/N19-1423} {{BERT}: Pre-training of
  deep bidirectional transformers for language understanding}.
\newblock In \emph{Proceedings of the 2019 Conference of the North {A}merican
  Chapter of the Association for Computational Linguistics: Human Language
  Technologies, Volume 1 (Long and Short Papers)}, pages 4171--4186,
  Minneapolis, Minnesota. Association for Computational Linguistics.

\bibitem[{Edunov et~al.(2020)Edunov, Ott, Ranzato, and
  Auli}]{edunov-etal-2020-evaluation}
Sergey Edunov, Myle Ott, Marc{'}Aurelio Ranzato, and Michael Auli. 2020.
\newblock \href {https://doi.org/10.18653/v1/2020.acl-main.253} {On the
  evaluation of machine translation systems trained with back-translation}.
\newblock In \emph{Proceedings of the 58th Annual Meeting of the Association
  for Computational Linguistics}, pages 2836--2846, Online. Association for
  Computational Linguistics.

\bibitem[{Eric et~al.(2020)Eric, Goel, Paul, Sethi, Agarwal, Gao, Kumar, Goyal,
  Ku, and Hakkani-Tur}]{eric-etal-2020-multiwoz}
Mihail Eric, Rahul Goel, Shachi Paul, Abhishek Sethi, Sanchit Agarwal, Shuyang
  Gao, Adarsh Kumar, Anuj Goyal, Peter Ku, and Dilek Hakkani-Tur. 2020.
\newblock \href {https://www.aclweb.org/anthology/2020.lrec-1.53} {{M}ulti{WOZ}
  2.1: A consolidated multi-domain dialogue dataset with state corrections and
  state tracking baselines}.
\newblock In \emph{Proceedings of the 12th Language Resources and Evaluation
  Conference}, pages 422--428, Marseille, France. European Language Resources
  Association.

\bibitem[{Glava{\v{s}} and Vuli{\'c}(2021)}]{glavas-vulic-2021-supervised}
Goran Glava{\v{s}} and Ivan Vuli{\'c}. 2021.
\newblock \href {https://www.aclweb.org/anthology/2021.eacl-main.270} {Is
  supervised syntactic parsing beneficial for language understanding tasks? an
  empirical investigation}.
\newblock In \emph{Proceedings of the 16th Conference of the European Chapter
  of the Association for Computational Linguistics: Main Volume}, pages
  3090--3104, Online. Association for Computational Linguistics.

\bibitem[{Gunasekara et~al.(2020)Gunasekara, Kim, D'Haro, Rastogi, Chen, Eric,
  Hedayatnia, Gopalakrishnan, Liu, Huang, Hakkani{-}T{\"{u}}r, Li, Zhu, Luo,
  Liden, Huang, Shayandeh, Liang, Peng, Zhang, Shukla, Huang, Gao, Mehri, Feng,
  Gordon, Alavi, Traum, Esk{\'{e}}nazi, Beirami, Cho, Crook, De, Geramifard,
  Kottur, Moon, Poddar, and Subba}]{DBLP:journals/corr/abs-2011-06486}
R.~Chulaka Gunasekara, Seokhwan Kim, Luis~Fernando D'Haro, Abhinav Rastogi,
  Yun{-}Nung Chen, Mihail Eric, Behnam Hedayatnia, Karthik Gopalakrishnan, Yang
  Liu, Chao{-}Wei Huang, Dilek Hakkani{-}T{\"{u}}r, Jinchao Li, Qi~Zhu,
  Lingxiao Luo, Lars Liden, Kaili Huang, Shahin Shayandeh, Runze Liang, Baolin
  Peng, Zheng Zhang, Swadheen Shukla, Minlie Huang, Jianfeng Gao, Shikib Mehri,
  Yulan Feng, Carla Gordon, Seyed~Hossein Alavi, David~R. Traum, Maxine
  Esk{\'{e}}nazi, Ahmad Beirami, Eunjoon Cho, Paul~A. Crook, Ankita De, Alborz
  Geramifard, Satwik Kottur, Seungwhan Moon, Shivani Poddar, and Rajen Subba.
  2020.
\newblock \href {http://arxiv.org/abs/2011.06486} {Overview of the ninth dialog
  system technology challenge: {DSTC9}}.
\newblock \emph{CoRR}, abs/2011.06486.

\bibitem[{Gururangan et~al.(2020)Gururangan, Marasovi{\'c}, Swayamdipta, Lo,
  Beltagy, Downey, and Smith}]{gururangan-etal-2020-dont}
Suchin Gururangan, Ana Marasovi{\'c}, Swabha Swayamdipta, Kyle Lo, Iz~Beltagy,
  Doug Downey, and Noah~A. Smith. 2020.
\newblock \href {https://doi.org/10.18653/v1/2020.acl-main.740} {Don{'}t stop
  pretraining: Adapt language models to domains and tasks}.
\newblock In \emph{Proceedings of the 58th Annual Meeting of the Association
  for Computational Linguistics}, pages 8342--8360, Online. Association for
  Computational Linguistics.

\bibitem[{Han and Eisenstein(2019)}]{han-eisenstein-2019-unsupervised}
Xiaochuang Han and Jacob Eisenstein. 2019.
\newblock \href {https://doi.org/10.18653/v1/D19-1433} {Unsupervised domain
  adaptation of contextualized embeddings for sequence labeling}.
\newblock In \emph{Proceedings of the 2019 Conference on Empirical Methods in
  Natural Language Processing and the 9th International Joint Conference on
  Natural Language Processing (EMNLP-IJCNLP)}, pages 4238--4248, Hong Kong,
  China. Association for Computational Linguistics.

\bibitem[{Howard and Ruder(2018)}]{howard-ruder-2018-universal}
Jeremy Howard and Sebastian Ruder. 2018.
\newblock \href {https://doi.org/10.18653/v1/P18-1031} {Universal language
  model fine-tuning for text classification}.
\newblock In \emph{Proceedings of the 56th Annual Meeting of the Association
  for Computational Linguistics (Volume 1: Long Papers)}, pages 328--339,
  Melbourne, Australia. Association for Computational Linguistics.

\bibitem[{Hu et~al.(2020)Hu, Ruder, Siddhant, Neubig, Firat, and
  Johnson}]{hu2020xtreme}
Junjie Hu, Sebastian Ruder, Aditya Siddhant, Graham Neubig, Orhan Firat, and
  Melvin Johnson. 2020.
\newblock \href {https://arxiv.org/pdf/2003.11080.pdf} {{XTREME}: A massively
  multilingual multi-task benchmark for evaluating cross-lingual
  generalisation}.
\newblock In \emph{International Conference on Machine Learning (ICML)}.

\bibitem[{Lee et~al.(2019)Lee, Lee, and Kim}]{lee-etal-2019-sumbt}
Hwaran Lee, Jinsik Lee, and Tae-Yoon Kim. 2019.
\newblock \href {https://doi.org/10.18653/v1/P19-1546} {{SUMBT}: Slot-utterance
  matching for universal and scalable belief tracking}.
\newblock In \emph{Proceedings of the 57th Annual Meeting of the Association
  for Computational Linguistics}, pages 5478--5483, Florence, Italy.
  Association for Computational Linguistics.

\bibitem[{Lin and Chen(2021)}]{lin2021empirical}
Yen-Ting Lin and Yun-Nung Chen. 2021.
\newblock \href {http://arxiv.org/abs/2101.11360} {An empirical study of
  cross-lingual transferability in generative dialogue state tracker}.

\bibitem[{Lison and Tiedemann(2016)}]{Lison2016OpenSubtitles2016EL}
P.~Lison and J.~Tiedemann. 2016.
\newblock Opensubtitles2016: Extracting large parallel corpora from movie and
  tv subtitles.
\newblock In \emph{LREC}.

\bibitem[{Liu et~al.(2020)Liu, Winata, Lin, Xu, and
  Fung}]{DBLP:conf/aaai/LiuWLXF20}
Zihan Liu, Genta~Indra Winata, Zhaojiang Lin, Peng Xu, and Pascale Fung. 2020.
\newblock \href {https://aaai.org/ojs/index.php/AAAI/article/view/6362}
  {Attention-informed mixed-language training for zero-shot cross-lingual
  task-oriented dialogue systems}.
\newblock In \emph{The Thirty-Fourth {AAAI} Conference on Artificial
  Intelligence, {AAAI} 2020, The Thirty-Second Innovative Applications of
  Artificial Intelligence Conference, {IAAI} 2020, The Tenth {AAAI} Symposium
  on Educational Advances in Artificial Intelligence, {EAAI} 2020, New York,
  NY, USA, February 7-12, 2020}, pages 8433--8440. {AAAI} Press.

\bibitem[{Mehri et~al.(2019)Mehri, Razumovskaia, Zhao, and
  Eskenazi}]{mehri-etal-2019-pretraining}
Shikib Mehri, Evgeniia Razumovskaia, Tiancheng Zhao, and Maxine Eskenazi. 2019.
\newblock \href {https://doi.org/10.18653/v1/P19-1373} {Pretraining methods for
  dialog context representation learning}.
\newblock In \emph{Proceedings of the 57th Annual Meeting of the Association
  for Computational Linguistics}, pages 3836--3845, Florence, Italy.
  Association for Computational Linguistics.

\bibitem[{Mrk{\v{s}}i{\'c} et~al.(2017{\natexlab{a}})Mrk{\v{s}}i{\'c},
  {\'O}~S{\'e}aghdha, Wen, Thomson, and Young}]{mrksic-etal-2017-neural}
Nikola Mrk{\v{s}}i{\'c}, Diarmuid {\'O}~S{\'e}aghdha, Tsung-Hsien Wen, Blaise
  Thomson, and Steve Young. 2017{\natexlab{a}}.
\newblock \href {https://doi.org/10.18653/v1/P17-1163} {Neural belief tracker:
  Data-driven dialogue state tracking}.
\newblock In \emph{Proceedings of the 55th Annual Meeting of the Association
  for Computational Linguistics (Volume 1: Long Papers)}, pages 1777--1788,
  Vancouver, Canada. Association for Computational Linguistics.

\bibitem[{Mrk{\v{s}}i{\'c} et~al.(2017{\natexlab{b}})Mrk{\v{s}}i{\'c},
  Vuli{\'c}, {\'O}~S{\'e}aghdha, Leviant, Reichart, Ga{\v{s}}i{\'c}, Korhonen,
  and Young}]{mrksic-etal-2017-semantic}
Nikola Mrk{\v{s}}i{\'c}, Ivan Vuli{\'c}, Diarmuid {\'O}~S{\'e}aghdha, Ira
  Leviant, Roi Reichart, Milica Ga{\v{s}}i{\'c}, Anna Korhonen, and Steve
  Young. 2017{\natexlab{b}}.
\newblock \href {https://doi.org/10.1162/tacl_a_00063} {Semantic specialization
  of distributional word vector spaces using monolingual and cross-lingual
  constraints}.
\newblock \emph{Transactions of the Association for Computational Linguistics},
  5:309--324.

\bibitem[{Phang et~al.(2020)Phang, Calixto, Htut, Pruksachatkun, Liu, Vania,
  Kann, and Bowman}]{phang-etal-2020-english}
Jason Phang, Iacer Calixto, Phu~Mon Htut, Yada Pruksachatkun, Haokun Liu, Clara
  Vania, Katharina Kann, and Samuel~R. Bowman. 2020.
\newblock \href {https://www.aclweb.org/anthology/2020.aacl-main.56} {{E}nglish
  intermediate-task training improves zero-shot cross-lingual transfer too}.
\newblock In \emph{Proceedings of the 1st Conference of the Asia-Pacific
  Chapter of the Association for Computational Linguistics and the 10th
  International Joint Conference on Natural Language Processing}, pages
  557--575, Suzhou, China. Association for Computational Linguistics.

\bibitem[{Phang et~al.(2019)Phang, Févry, and Bowman}]{phang2019sentence}
Jason Phang, Thibault Févry, and Samuel~R. Bowman. 2019.
\newblock \href {http://arxiv.org/abs/1811.01088} {Sentence encoders on stilts:
  Supplementary training on intermediate labeled-data tasks}.

\bibitem[{Pruksachatkun et~al.(2020)Pruksachatkun, Phang, Liu, Htut, Zhang,
  Pang, Vania, Kann, and Bowman}]{pruksachatkun-etal-2020-intermediate}
Yada Pruksachatkun, Jason Phang, Haokun Liu, Phu~Mon Htut, Xiaoyi Zhang,
  Richard~Yuanzhe Pang, Clara Vania, Katharina Kann, and Samuel~R. Bowman.
  2020.
\newblock \href {https://doi.org/10.18653/v1/2020.acl-main.467}
  {Intermediate-task transfer learning with pretrained language models: When
  and why does it work?}
\newblock In \emph{Proceedings of the 58th Annual Meeting of the Association
  for Computational Linguistics}, pages 5231--5247, Online. Association for
  Computational Linguistics.

\bibitem[{Qin et~al.(2020)Qin, Ni, Zhang, and Che}]{DBLP:conf/ijcai/QinN0C20}
Libo Qin, Minheng Ni, Yue Zhang, and Wanxiang Che. 2020.
\newblock \href {https://doi.org/10.24963/ijcai.2020/533} {Cosda-ml:
  Multi-lingual code-switching data augmentation for zero-shot cross-lingual
  {NLP}}.
\newblock In \emph{Proceedings of the Twenty-Ninth International Joint
  Conference on Artificial Intelligence, {IJCAI} 2020}, pages 3853--3860.
  ijcai.org.

\bibitem[{Qiu et~al.(2020)Qiu, Sun, Xu, Shao, Dai, and
  Huang}]{qiu2020pretrained}
Xipeng Qiu, Tianxiang Sun, Yige Xu, Yunfan Shao, Ning Dai, and Xuanjing Huang.
  2020.
\newblock \href {http://arxiv.org/abs/2003.08271} {Pre-trained models for
  natural language processing: A survey}.

\bibitem[{Ren et~al.(2018)Ren, Xie, Chen, and Yu}]{ren-etal-2018-towards}
Liliang Ren, Kaige Xie, Lu~Chen, and Kai Yu. 2018.
\newblock \href {https://doi.org/10.18653/v1/D18-1299} {Towards universal
  dialogue state tracking}.
\newblock In \emph{Proceedings of the 2018 Conference on Empirical Methods in
  Natural Language Processing}, pages 2780--2786, Brussels, Belgium.
  Association for Computational Linguistics.

\bibitem[{Schuster et~al.(2019)Schuster, Gupta, Shah, and
  Lewis}]{schuster-etal-2019-cross-lingual}
Sebastian Schuster, Sonal Gupta, Rushin Shah, and Mike Lewis. 2019.
\newblock \href {https://doi.org/10.18653/v1/N19-1380} {Cross-lingual transfer
  learning for multilingual task oriented dialog}.
\newblock In \emph{Proceedings of the 2019 Conference of the North {A}merican
  Chapter of the Association for Computational Linguistics: Human Language
  Technologies, Volume 1 (Long and Short Papers)}, pages 3795--3805,
  Minneapolis, Minnesota. Association for Computational Linguistics.

\bibitem[{Taylor(1953)}]{doi:10.1177/107769905303000401}
Wilson~L. Taylor. 1953.
\newblock \href {https://doi.org/10.1177/107769905303000401} {“cloze
  procedure”: A new tool for measuring readability}.
\newblock \emph{Journalism Quarterly}, 30(4):415--433.

\bibitem[{Vaswani et~al.(2017)Vaswani, Shazeer, Parmar, Uszkoreit, Jones,
  Gomez, Kaiser, and Polosukhin}]{Vaswani2017-sz}
Ashish Vaswani, Noam Shazeer, Niki Parmar, Jakob Uszkoreit, Llion Jones,
  Aidan~N Gomez, {\L}~Ukasz Kaiser, and Illia Polosukhin. 2017.
\newblock Attention is all you need.
\newblock In I~Guyon, U~V Luxburg, S~Bengio, H~Wallach, R~Fergus,
  S~Vishwanathan, and R~Garnett, editors, \emph{Advances in Neural Information
  Processing Systems 30}, pages 5998--6008. Curran Associates, Inc.

\bibitem[{Wen et~al.(2017)Wen, Vandyke, Mrk{\v{s}}i{\'c}, Ga{\v{s}}i{\'c},
  Rojas-Barahona, Su, Ultes, and Young}]{wen-etal-2017-network}
Tsung-Hsien Wen, David Vandyke, Nikola Mrk{\v{s}}i{\'c}, Milica
  Ga{\v{s}}i{\'c}, Lina~M. Rojas-Barahona, Pei-Hao Su, Stefan Ultes, and Steve
  Young. 2017.
\newblock \href {https://www.aclweb.org/anthology/E17-1042} {A network-based
  end-to-end trainable task-oriented dialogue system}.
\newblock In \emph{Proceedings of the 15th Conference of the {E}uropean Chapter
  of the Association for Computational Linguistics: Volume 1, Long Papers},
  pages 438--449, Valencia, Spain. Association for Computational Linguistics.

\bibitem[{Wolf et~al.(2020)Wolf, Debut, Sanh, Chaumond, Delangue, Moi, Cistac,
  Rault, Louf, Funtowicz, Davison, Shleifer, von Platen, Ma, Jernite, Plu, Xu,
  Le~Scao, Gugger, Drame, Lhoest, and Rush}]{wolf-etal-2020-transformers}
Thomas Wolf, Lysandre Debut, Victor Sanh, Julien Chaumond, Clement Delangue,
  Anthony Moi, Pierric Cistac, Tim Rault, Remi Louf, Morgan Funtowicz, Joe
  Davison, Sam Shleifer, Patrick von Platen, Clara Ma, Yacine Jernite, Julien
  Plu, Canwen Xu, Teven Le~Scao, Sylvain Gugger, Mariama Drame, Quentin Lhoest,
  and Alexander Rush. 2020.
\newblock \href {https://doi.org/10.18653/v1/2020.emnlp-demos.6} {Transformers:
  State-of-the-art natural language processing}.
\newblock In \emph{Proceedings of the 2020 Conference on Empirical Methods in
  Natural Language Processing: System Demonstrations}, pages 38--45, Online.
  Association for Computational Linguistics.

\bibitem[{Wolf et~al.(2019)Wolf, Sanh, Chaumond, and
  Delangue}]{wolf2019transfertransfo}
Thomas Wolf, Victor Sanh, Julien Chaumond, and Clement Delangue. 2019.
\newblock \href {http://arxiv.org/abs/1901.08149} {Transfertransfo: A transfer
  learning approach for neural network based conversational agents}.

\bibitem[{Zhu et~al.(2020{\natexlab{a}})Zhu, Huang, Zhang, Zhu, and
  Huang}]{DBLP:journals/tacl/ZhuHZZH20}
Qi~Zhu, Kaili Huang, Zheng Zhang, Xiaoyan Zhu, and Minlie Huang.
  2020{\natexlab{a}}.
\newblock \href {https://transacl.org/ojs/index.php/tacl/article/view/1901}
  {Crosswoz: {A} large-scale chinese cross-domain task-oriented dialogue
  dataset}.
\newblock \emph{Trans. Assoc. Comput. Linguistics}, 8:281--295.

\bibitem[{Zhu et~al.(2020{\natexlab{b}})Zhu, Zhang, Fang, Li, Takanobu, Li,
  Peng, Gao, Zhu, and Huang}]{zhu-etal-2020-convlab}
Qi~Zhu, Zheng Zhang, Yan Fang, Xiang Li, Ryuichi Takanobu, Jinchao Li, Baolin
  Peng, Jianfeng Gao, Xiaoyan Zhu, and Minlie Huang. 2020{\natexlab{b}}.
\newblock \href {https://doi.org/10.18653/v1/2020.acl-demos.19} {{C}onv{L}ab-2:
  An open-source toolkit for building, evaluating, and diagnosing dialogue
  systems}.
\newblock In \emph{Proceedings of the 58th Annual Meeting of the Association
  for Computational Linguistics: System Demonstrations}, pages 142--149,
  Online. Association for Computational Linguistics.

\end{thebibliography}
